\documentclass[11pt,letterpaper]{article}
\usepackage[T1]{fontenc}
\usepackage[utf8]{inputenc}
\usepackage{lmodern}
\usepackage[left=1.05in,right=1.05in,top=1in,bottom=1.1in,includefoot]{geometry}

\usepackage{amsmath,amssymb,amsthm,booktabs,tabularx,array,graphicx,microtype}
\usepackage{tikz,flafter,float,placeins}
\usetikzlibrary{arrows.meta,positioning}
\usepackage[round,authoryear]{natbib}
\usepackage{hyperref,url}
\makeatletter
\g@addto@macro\UrlBreaks{\do\a\do\b\do\c\do\d\do\e\do\f\do\g\do\h\do\i\do\j\do\k\do\l\do\m\do\n\do\o\do\p\do\q\do\r\do\s\do\t\do\u\do\v\do\w\do\x\do\y\do\z}
\makeatother
\microtypesetup{protrusion=false}
\newcolumntype{Y}{>{\raggedright\arraybackslash}X}
\newcommand{\Qhat}{\widehat Q_{E,B}}
\newcommand{\asctagent}{\textbf{ASCT-AgentUCT}}
\floatstyle{ruled}
\newfloat{algorithm}{tbp}{loa}
\floatname{algorithm}{Algorithm}
\title{ASCT: Attentive Search over Counterfactual Trees\\for Credit Assignment in\\Agentic Reinforcement Learning}
\author{\normalsize\begin{tabular}{c}
Yang Li\textsuperscript{1}\quad Jinhan Yang\textsuperscript{2}\quad hai liu\textsuperscript{3}\quad Di Wan\\[3pt]
Xiyu Chen\textsuperscript{1}\quad Zongsi Xu\textsuperscript{1}\quad Tuo Zhou\textsuperscript{1}\quad Sheng Zhong\textsuperscript{1}\\[3pt]
Sergey Volkov\textsuperscript{1}\quad Ye Luo\textsuperscript{1,*}\quad Hao Sun\textsuperscript{5,*}
\end{tabular}\\[8pt]
\small\begin{tabular}{c}
\textsuperscript{1}University of Hong Kong (hku.hk)\\[2pt]
\textsuperscript{2}The Chinese University of Hong Kong (cuhk.edu.hk)\\[2pt]
\textsuperscript{3}Jiangxi Science and Technology Normal University (jxstnu.edu.cn)\\[2pt]
\textsuperscript{5}Shenzhen University (szu.edu.cn)
\end{tabular}\\[6pt]
\small\textsuperscript{*}Corresponding authors: Ye Luo and Hao Sun.}
\date{}
\hypersetup{hidelinks,pdftitle={ASCT: Attentive Search over Counterfactual Trees for Credit Assignment in Agentic Reinforcement Learning},pdfauthor={Yang Li; Jinhan Yang; hai liu; Di Wan; Xiyu Chen; Zongsi Xu; Tuo Zhou; Sheng Zhong; Sergey Volkov; Ye Luo; Hao Sun}}
\begin{document}
	\maketitle
	
	\begin{abstract}
		Terminal utility evaluates a complete agentic workflow, but learning requires credit for the decisions within it. We introduce Attentive Search over Counterfactual Trees (ASCT), a framework that turns training-time multi-step search into local action credit. At actor-visited states, an auxiliary tree evaluates alternative legal actions from the same recoverable prefix. Its action-value table is centered by the frozen actor's probabilities and supplies credit for PPO on actor-sampled trajectories. This protocol connects counterfactual evaluation to policy learning while deploying the actor alone. Uniform, UCT, and cost-aware AgentUCT instantiate the framework. On HotpotQA agentic retrieval-augmented generation, all three improve mean held-out utility over trajectory-return PPO and workflow-adapted VinePPO. Across three seeds, ASCT-AgentUCT reaches 0.6187 utility versus 0.5939 for VinePPO, with gains in answer F1 and execution cost, and uses 50.3\% fewer recorded auxiliary Qwen tokens. Transfer and component-description studies examine the learned policies beyond the training setting. 
	\end{abstract}
	
	\section{Introduction}
	\label{sec:intro}
	A workflow actor, the trainable large language model (LLM) policy in this paper, chooses components, gathers evidence, and decides when to stop. Terminal utility combines answer quality and execution cost, but a successful workflow can still contain a poor local choice. Learning requires credit that connects each decision to its downstream consequences.
	
	VinePPO estimates state values from Monte Carlo continuations for step-wise credit \citep{kazemnejad2025vineppo}; branching learners derive updates from collections of search trajectories. We study an action-resolved interface: compare executable alternatives at states the actor visits, then use those comparisons to train its own sampled decisions.
	
	Recoverable prefixes make this possible. From the same observations, an evaluator can execute alternative legal actions through later workflow decisions to terminal outcomes. Their value depends on these continuations: different retrieved evidence can change whether another retrieval round is useful. Finite action sets permit root coverage, while shared prefixes permit execution reuse within a limited auxiliary budget.
	
	We introduce Attentive Search over Counterfactual Trees (ASCT). To our knowledge, ASCT is the first agentic RL framework to use multi-step counterfactual tree search as an auxiliary legal-action value estimator for actor-centered credit on separately sampled actor trajectories. Search constructs a local Q table; frozen actor probabilities center its values; PPO updates the actor's sampled decisions with their recorded behavior probabilities. Deployment uses the actor alone.
	
	All three evaluators---Uniform, UCT, and AgentUCT---improve mean held-out utility over trajectory-return PPO and workflow-adapted VinePPO. Across three seeds, ASCT-AgentUCT reaches 0.6187 versus 0.5939 for VinePPO, combining higher answer F1 with lower execution cost and using 50.3\% fewer recorded auxiliary Qwen tokens. Transfer and description-controlled experiments examine the learned policies beyond the training setting.
	
	ASCT connects: \textbf{multi-step action comparison} from the same recoverable state; \textbf{separate evaluation allocation and behavior sampling}, linked by actor-centered credit and consistent PPO records; and \textbf{training-time information acquisition amortized into an actor-only policy}. Uniform, UCT, AgentUCT, and WTB provide inherited search and execution components \citep{li2026agentuct}. We evaluate the resulting learning protocol, acquisition cost, and transfer.	
	
	\section{Related Work}
	\label{sec:related}
	\paragraph{Auxiliary evaluation and credit.}
	PPO updates sampled decisions using advantage estimates \citep{schulman2017ppo}. VinePPO supplies step credit from Monte Carlo state values \citep{kazemnejad2025vineppo}. It is our closest estimator comparison: both credit the main actor trajectory, allowing us to compare state-value credit with ASCT's action table. COMA uses a policy-weighted counterfactual baseline with a centralized critic \citep{foerster2018coma}; ASCT estimates values through workflow execution.
	
	\paragraph{Tree-based RL for agentic decisions.}
	BPO uses sibling-relative credit at high-entropy branch states; Tree-GRPO uses intra- and inter-tree advantages; TreePS-RAG derives process supervision from descendants \citep{he2026bpo,ji2026treegrpo,zhang2026treeps}. ATPO combines dialogue branching with critic-based credit, and AT$^2$PO uses turn-level tree credit \citep{cao2026atpo,zong2026at2po}. These learners optimize branched rollout collections. ASCT's tree estimates legal-action values at separately sampled actor states; only actor-trajectory decisions enter PPO. Its multi-step branches execute evidence-dependent component choices, linking workflow consequences to this distinct learning interface.
	
	\paragraph{Search allocation and execution reuse.}
	UCT allocates trials through reward and uncertainty \citep{auer2002ucb,kocsis2006uct}. LATS and related methods search for deployed actions \citep{zhou2024lats,koh2024search}. We use AgentUCT's reuse-aware allocation and WTB execution infrastructure \citep{li2026agentuct} to acquire training credit, then assess both acquisition cost and the learned actor.
	
	\section{Problem Formulation}
	\label{sec:background}
	A state $s_t$ contains a task, its executed workflow prefix, and observations. The actor $\pi_\theta(a\mid s_t)$ chooses from a finite legal set $\mathcal A(s_t)$. A terminal workflow $w$ receives
	\begin{equation}
		U(w)=P(w)-\lambda C_{\mathrm{exec}}(w)/C_0,
		\label{eq:utility}
	\end{equation}
	where $P$ is answer quality and $C_{\mathrm{exec}}$ is execution cost. The learning objective is expected terminal utility. Recoverable prefixes permit alternative continuations at actor-visited states. We hold the PPO learner fixed to compare credit estimators, with $B$ terminal trials per evaluated state. Equal logical budgets can require different physical work.
	
	Auxiliary work is logged separately from workflow execution utility; RQ2 defines the cost-adjusted metrics.
	
	\section{ASCT}
	\label{sec:method}
\begin{figure}[!htb]
\centering
\includegraphics[width=\linewidth]{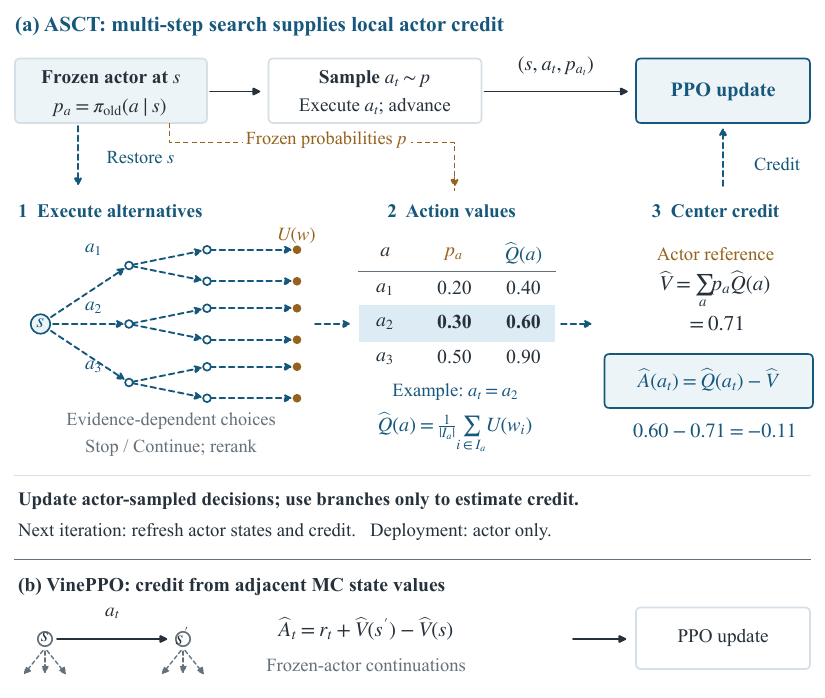}
\caption{ASCT action-value credit and VinePPO state-value credit share actor-trajectory PPO updates. Solid arrows carry behavior records; dashed arrows carry auxiliary information. Frozen probabilities center ASCT's action table. Values and tree geometry are illustrative; subscripts are suppressed.}
\label{fig:training_interface}
\end{figure}

	\textbf{An agentic decision example.} At a retriever choice, the actor may favor a familiar retriever. Another legal retriever could expose evidence that makes an extra retrieval round useful; its value emerges after continuation, stopping, and reranking. ASCT restores the shared prefix and evaluates these multi-step outcomes. The actor still executes its own sampled action; search supplies the comparison used to train that decision (Figure~\ref{fig:training_interface}).
	
	\textbf{Separate behavior from evaluation.} At visited state $s$, the frozen actor computes $p_a=\pi_{\mathrm{old}}(a\mid s)$ and samples $a_t$. An independent random stream evaluates alternatives from recoverable copies of $s$. Search allocation can change while the actor advances with $a_t$.
	
	\textbf{Retain multi-step action values.} Tree nodes are workflow prefixes and edges are legal actions. Trials execute selected paths to completion and back up $U$. Root coverage evaluates every action before repeat allocation ($B\geq|\mathcal A(s)|$). Each value averages the returns of its root-action trials:
	\begin{equation}
		\Qhat(s,a)=\frac{1}{|\mathcal I_{E,B}(s,a)|}\sum_{i\in\mathcal I_{E,B}(s,a)}U(w_i).
		\label{eq:qtable}
	\end{equation}
	Here $\mathcal I_{E,B}(s,a)$ indexes trials starting with $a$; evaluator $E$ and budget $B$ determine their suffixes.

	\textbf{Credit the sampled action relative to the actor.}
	\begin{align}
		\widehat V_E(s)&=\sum_{a\in\mathcal A(s)}p_a\Qhat(s,a),\label{eq:baseline}\\
		\widehat A_E(s,a_t)&=\Qhat(s,a_t)-\widehat V_E(s).\label{eq:advantage}
	\end{align}
	For fixed $\Qhat$, let $L_s(\theta)=\sum_a\pi_\theta(a\mid s)\Qhat(s,a)$. At $\pi_\theta=\pi_{\mathrm{old}}$, its derivative with respect to softmax logit $z_a$ is $\partial L_s/\partial z_a=p_a\widehat A_E(s,a)$. Thus, actor-centered credit provides a local policy-gradient signal before normalization and clipping.  
	
	\textbf{Reuse execution and update online.} Uniform balances visits; UCT uses utility and exploration; AgentUCT also accounts for predicted uncached cost. WTB restores shared prefixes and executes uncached suffixes for all evaluators and VinePPO. Each actor decision contributes $(s,a_t,p_{a_t},\widehat A_E(s,a_t))$. After an iteration, credit is standardized and PPO uses $\rho_\theta=\pi_\theta(a_t\mid s)/p_{a_t}$ \citep{schulman2017ppo}. One frozen actor defines sampling, baseline weights, and the ratio denominator; auxiliary branches supply credit without entering the update trajectories. The next iteration refreshes visited states and credit with the updated actor. Deployment runs the actor alone. Algorithm~\ref{alg:asct} and Appendix~\ref{app:config} specify the full procedure and objective; Appendix~\ref{app:paired} illustrates evaluator rules.
	
	\section{Experiments}
	\label{sec:experiments}
	Four RQs evaluate final-policy effectiveness, auxiliary cost, cross-dataset transfer, and description-guided choice. Methods share the workflow and learning protocol. Section~\ref{sec:discussion} relates these policy results to supplementary evaluation controls.
	
	\subsection{Agentic workflow and RAGSpace}
	\label{sec:ragspace}
	We instantiate the workflow in RAGSpace on HotpotQA questions and supplied candidate passages \citep{yang2018hotpotqa,li2026agentuct}. The actor successively chooses query formulation, retrieval, evidence selection, continuation, and reranking (Table~\ref{tab:grammar}). Retrieval and generation consume passage text; answer annotations score completed workflows. The shared generator and retrieval components remain frozen.
	
	The query choices are the original question, Query2Doc-style expansion \citep{wang2023query2doc}, and two-query decomposition. Retrievers are BM25 \citep{robertson2009bm25}, E5-base-v2 \citep{wang2022e5}, BGE-M3 hybrid \citep{chen2024m3}, Qwen3-Embedding-0.6B \citep{zhang2025qwenembedding}, and reciprocal-rank fusion of their rankings \citep{cormack2009rrf}. Passage selection uses relevance or maximal marginal relevance \citep{carbonell1998mmr}; optional reranking uses Qwen3-Reranker-0.6B \citep{zhang2025qwenembedding}.
	
	\begin{table}[!htb]
		\centering
		\small
		\caption{Executable RAG decisions. A workflow uses up to three retrieval rounds. Choices within a row are legal alternatives at that stage.}
		\label{tab:grammar}
		\begin{tabularx}{\linewidth}{@{}>{\raggedright\arraybackslash}p{0.30\linewidth} Y@{}}
			\toprule
			\textbf{Stage} & \textbf{Actions} \\
			\midrule
			Query formulation & Original question; Query2Doc; two-query decomposition \\
			Retriever & BM25; E5 dense; BGE-M3 hybrid; Qwen3 embedding; reciprocal-rank fusion \\
			Passage selection & Relevance ranking; maximal marginal relevance \\
			Retrieval width & 3 or 6 passages \\
			Retrieval control & Stop; continue while fewer than three rounds have executed \\
			Reranking & Preserve order; Qwen3 reranker \\
			Answer context & Top 2 or top 4 evidence passages \\
			\bottomrule
		\end{tabularx}
	\end{table}
	
	\textbf{Evidence-dependent control.} After retrieval, Stop ends acquisition; Continue refines the query from retrieved evidence, repeats the chosen retrieval rule, and merges passages. The actor then chooses again. At round three only Stop is legal; reranking and context-width choices finish the workflow. Thus earlier choices change later evidence and the value of continuing. Prompts and deterministic refinement are fixed across methods (Appendix~\ref{app:prompts}).
	
	\textbf{Actor and observations.} The planner sees the question, stage, selected components, round, active queries, evidence titles, and legal labels. It scores legal action responses and normalizes their probabilities; training samples actions and held-out evaluation is greedy. The actor is Qwen3-4B-Instruct-2507 \citep{qwen2025instruct2507} with LoRA \citep{hu2022lora}; query and answer generation use the frozen backbone with its adapter disabled. Learning changes workflow decisions while keeping component implementations fixed.
	
	\textbf{Protocol and comparisons.} All methods share the actor, workflow, splits, and PPO configuration (batch size eight; two update epochs per iteration). We train for three iterations on 2,000 questions and evaluate final checkpoints on 400 test questions over seeds 11, 23, and 37. Training uses the repository F1 scorer; endpoint reporting uses official F1, with $\lambda=0.1$ and $C_0=4096$. VinePPO estimates actor-continuation state values for one-step, unit-discount credit; PPO uses terminal trajectory credit. ASCT-Uniform, UCT, and AgentUCT vary the evaluator. VinePPO and ASCT share $B=12$ and WTB reuse. Logical budgets and realized work are reported separately. We set $c_{\mathrm{exp}}=1.4$ and $c_{\mathrm{tok}}=\beta=10^{-4}$; Appendix~\ref{app:config} gives the full configuration.
	
	\subsection{RQ1: Does counterfactual credit improve the final policy?}
	\label{sec:rq1}
	\begin{table}[!htb]
		\centering
		\small
		\caption{HotpotQA held-out results of the final policies. Workflow utility is the mean over test questions, using official F1 and execution words. Trained rows are mean \(\pm\) sample SD over three seeds.}
		\label{tab:main}
		\begin{tabularx}{\linewidth}{@{}Y r r r@{}}
			\toprule
			\textbf{Method} & \textbf{Workflow utility} & \textbf{Answer F1} & \textbf{Execution words} \\
			\midrule
			Base & 0.5648 & 0.6954 & 5,348.7 \\
			PPO & 0.5651 \(\pm\) 0.0021 & 0.6913 \(\pm\) 0.0021 & 5,170.4 \(\pm\) 76.8 \\
			VinePPO & 0.5939 \(\pm\) 0.0062 & 0.7115 \(\pm\) 0.0026 & 4,817.7 \(\pm\) 179.7 \\
			\textbf{\boldmath ASCT-Uniform} & \textbf{\boldmath 0.6146 \(\pm\) 0.0191} & \textbf{\boldmath 0.7316 \(\pm\) 0.0168} & \textbf{\boldmath 4,792.2 \(\pm\) 151.4} \\
			\textbf{\boldmath ASCT-UCT} & \textbf{\boldmath 0.6075 \(\pm\) 0.0241} & \textbf{\boldmath 0.7257 \(\pm\) 0.0260} & \textbf{\boldmath 4,841.2 \(\pm\) 112.8} \\
			\asctagent{} & \textbf{\boldmath 0.6187 \(\pm\) 0.0076} & \textbf{\boldmath 0.7266 \(\pm\) 0.0126} & \textbf{\boldmath 4,420.8 \(\pm\) 213.3} \\
			\bottomrule
		\end{tabularx}
	\end{table}
	
	\begin{figure}[!htb]
		\centering
		\includegraphics[width=.85\linewidth]{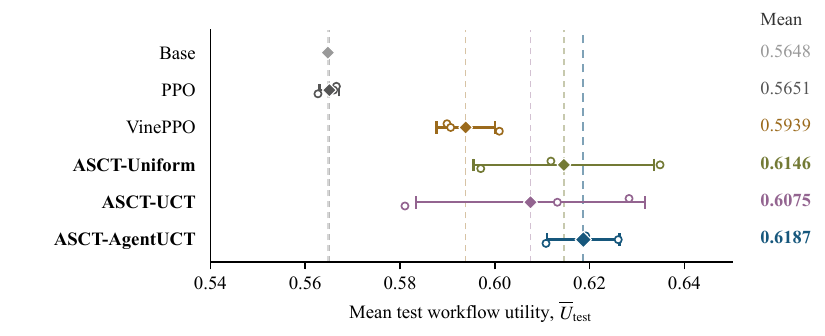}
		\caption{HotpotQA test utility. Diamonds and dashed guides mark means; bars show $\pm 1$ sample SD over three seeds; open circles show seeds. Base is fixed; ASCT labels are bold.}
		\label{fig:utility}
	\end{figure}
	
	All three ASCT evaluators improve mean held-out utility over VinePPO and trajectory-return PPO (Table~\ref{tab:main}, Figure~\ref{fig:utility}). ASCT-AgentUCT reaches 0.6187, a gain of 0.0248 over VinePPO: 0.0151 comes from higher answer F1 and 0.0097 from lower execution cost. Execution words decrease by 8.24\%. The paired utility differences are positive at all three seeds: 0.0293, 0.0354, and 0.0098.
	
	ASCT-Uniform reaches utility 0.6146 and the highest mean F1, 0.7316, supporting the interface with balanced evaluation. The smaller AgentUCT--Uniform difference, 0.0041 $\pm$ 0.0116, changes sign across seeds; acquisition cost also matters to evaluator choice.
	
\begin{figure}[!htb]
		\centering
		\includegraphics[width=\linewidth]{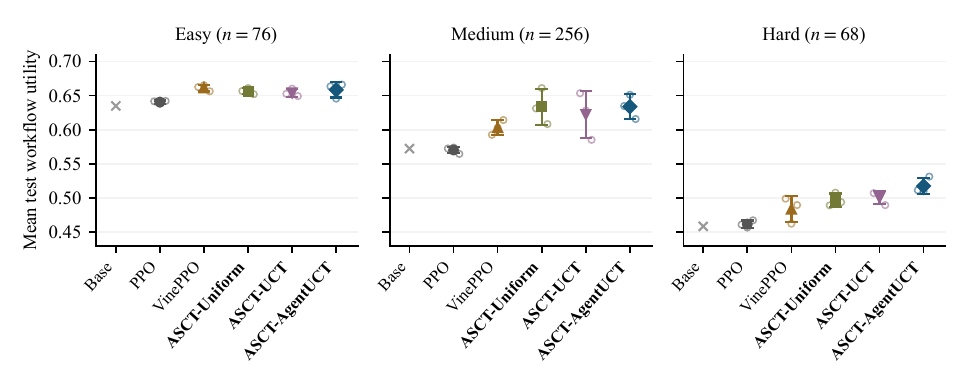}
		\caption{Test utility by difficulty: mean $\pm$ sample SD over three seeds, with open seed markers. Base is fixed. Panels share the utility scale and method order.}
		\label{fig:difficulty}
	\end{figure}

	\textbf{Where the gains occur.} All ASCT policies improve medium- and hard-question utility over VinePPO and PPO (Figure~\ref{fig:difficulty}). AgentUCT gains 0.0307 and 0.0339 over VinePPO, respectively; the easy group is closely matched, with VinePPO slightly higher.

	\subsection{RQ2: What does auxiliary evaluation cost, and when is it valuable?}
	\label{sec:rq3}
	Auxiliary tokens are $T_{\mathrm{aux}}=T_{\mathrm{env}}+T_{\mathrm{actor,aux}}$: recorded Qwen environment calls and auxiliary actor scoring after reuse; E5, BGE-M3, and index construction are outside this token sum. For $K$ logical terminal trials and an assumed $N$ subsequent policy uses, define
	\begin{align}
		\overline J_{\mathrm{search}}&=\bigl[\textstyle\sum_{i=1}^{K}U(w_i)-\beta T_{\mathrm{aux}}\bigr]/K,\label{eq:jsearch}\\
		J_{\mathrm{deploy}}(N)&=\overline U_{\mathrm{test}}-\beta T_{\mathrm{aux}}/N.\label{eq:jdeploy}
	\end{align}
	The first measures acquired outcomes after their cost; the second projects auxiliary-cost amortization, assuming held-out utility represents future use. Search backup and policy credit use $U$. Main-trajectory collection and optimizer scoring have separate counters. The common token weight $\beta$ defines an accounting convention; GPU time or API price requires model-specific calibration (Appendix~\ref{app:tokens}).

	\textbf{Auxiliary tokens for the recorded Qwen operations.} Table~\ref{tab:token_ledger} separates environment execution from actor scoring. VinePPO uses 137.09 million environment tokens and 251.29 million tokens to score auxiliary actor continuations, totaling 388.37 million. ASCT obtains continuation decisions from its search rules and incurs no auxiliary actor-scoring calls. AgentUCT uses 193.09 million tokens in this auxiliary ledger, 50.3\% below VinePPO and 2.02--2.28\% below the other ASCT evaluators. The within-ASCT reductions occur in all three seeds. The 50.3\% difference chiefly reflects avoided actor-continuation scoring; VinePPO uses fewer environment tokens. The cost-aware evaluator contributes the smaller within-ASCT reduction.
	
	\begin{table}[!htb]
\centering
\small
\setlength{\tabcolsep}{3pt}
\caption{Combined token ledger for the listed Qwen model operations across three training iterations, in millions (mean $\pm$ sample SD over three seeds). The first row sums auxiliary environment execution and auxiliary actor scoring within each seed. Indented rows are components of their section subtotal.}
\label{tab:token_ledger}
\begin{tabularx}{\linewidth}{@{}Y r r r r@{}}
\toprule
\textbf{Token count (M)} & \textbf{VinePPO} & \shortstack{\textbf{ASCT-}\\\textbf{Uniform}} & \shortstack{\textbf{ASCT-}\\\textbf{UCT}} & \shortstack{\textbf{ASCT-}\\\textbf{AgentUCT}} \\
\midrule
\textbf{Auxiliary total} & \textbf{\boldmath 388.374 $\pm$ 4.694} & \textbf{\boldmath 197.077 $\pm$ 0.024} & \textbf{\boldmath 197.588 $\pm$ 0.068} & \textbf{\boldmath 193.088 $\pm$ 0.404} \\
\midrule
\multicolumn{5}{@{}l}{\textit{Auxiliary environment execution}} \\
Environment subtotal $T_{\mathrm{env}}$ & 137.088 $\pm$ 1.231 & 197.077 $\pm$ 0.024 & 197.588 $\pm$ 0.068 & 193.088 $\pm$ 0.404 \\
\quad Generator input (4B) & 89.972 $\pm$ 0.278 & 139.452 $\pm$ 0.162 & 139.613 $\pm$ 0.193 & 137.776 $\pm$ 0.174 \\
\quad Generator output (4B) & 3.590 $\pm$ 0.009 & 5.874 $\pm$ 0.003 & 5.883 $\pm$ 0.002 & 5.833 $\pm$ 0.003 \\
\quad Query embedding (0.6B) & 0.611 $\pm$ 0.012 & 0.689 $\pm$ 0.002 & 0.690 $\pm$ 0.001 & 0.689 $\pm$ 0.002 \\
\quad Reranker input (0.6B) & 42.916 $\pm$ 1.452 & 51.062 $\pm$ 0.152 & 51.401 $\pm$ 0.148 & 48.789 $\pm$ 0.227 \\
\midrule
\multicolumn{5}{@{}l}{\textit{Actor scoring (4B)}} \\
Actor-scoring subtotal & 305.234 $\pm$ 3.515 & 76.325 $\pm$ 0.038 & 76.208 $\pm$ 0.051 & 76.115 $\pm$ 0.148 \\
\quad Auxiliary continuations & 251.286 $\pm$ 3.472 & 0.000 $\pm$ 0.000 & 0.000 $\pm$ 0.000 & 0.000 $\pm$ 0.000 \\
\quad Main-trajectory collection & 3.151 $\pm$ 0.013 & 25.442 $\pm$ 0.013 & 25.403 $\pm$ 0.017 & 25.372 $\pm$ 0.049 \\
\quad Optimizer forwards & 50.796 $\pm$ 0.044 & 50.883 $\pm$ 0.025 & 50.805 $\pm$ 0.034 & 50.744 $\pm$ 0.098 \\
\bottomrule
\end{tabularx}
\par\smallskip\raggedright\footnotesize The main $J$ metrics use the auxiliary total: environment subtotal plus auxiliary-continuation scoring. Main-trajectory collection and optimizer forwards contribute to the actor-scoring subtotal.
\end{table}

	VinePPO reuses frozen action probabilities within each question; scoring is charged once to the first requesting path.

\begin{figure}[!htb]
		\begin{minipage}[t]{.485\linewidth}
			\centering
			\includegraphics[width=\linewidth]{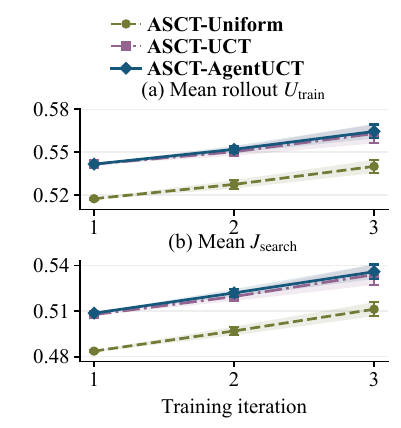}
			\caption{Per-iteration auxiliary evaluation within ASCT: mean rollout $U_{\mathrm{train}}$ and $J_{\mathrm{search}}$ (2,000 questions; $B=12$). Utility uses the training scorer. Bands and bars: $\pm$ sample SD across three seeds.}
			\label{fig:training_cost}
		\end{minipage}\hfill
		\begin{minipage}[t]{.485\linewidth}
			\centering
			\includegraphics[width=\linewidth]{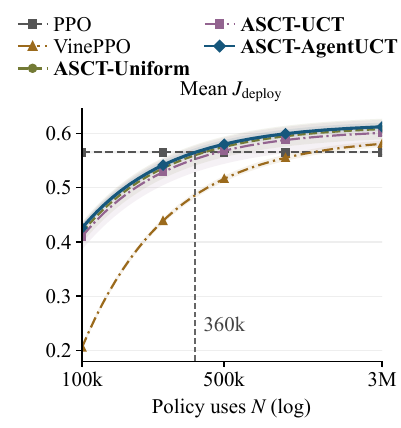}
			\caption{Amortized value across assumed policy uses $N$ (Equation~\ref{eq:jdeploy}; $\beta=10^{-4}$). Bands: $\pm$ sample SD across three seed projections. The guide marks AgentUCT's mean crossing with PPO. Both $J$ metrics charge the recorded auxiliary ledger.}
			\label{fig:amortization}
		\end{minipage}
	\end{figure}

	\textbf{Cost-adjusted evaluation value.} Table~\ref{tab:j_primary} aggregates rollout utility, auxiliary tokens, and $\overline J_{\mathrm{search}}$ over complete runs. Figure~\ref{fig:training_cost} shows per-iteration outcomes within ASCT. AgentUCT has the highest mean $J_{\mathrm{search}}$ in each iteration and the lowest full-run auxiliary cost. Both use environment execution plus auxiliary actor scoring.
	
	\begin{table}[!htb]
\centering\small
\setlength{\tabcolsep}{3pt}
\caption{Mean auxiliary rollout utility, recorded auxiliary token cost, and $\overline J_{\mathrm{search}}$ over complete training runs. Rollout utility uses the training scorer; tokens include the listed Qwen environment operations and auxiliary actor scoring. Values are mean $\pm$ sample SD across three seeds.}
\label{tab:j_primary}
\begin{tabularx}{\linewidth}{@{}Y r r r@{}}
\toprule
Evaluator & Mean rollout $U$ & Auxiliary tokens (M) & $\overline J_{\mathrm{search}}$ \\
\midrule
VinePPO & 0.5507 $\pm$ 0.0027 & 388.374 $\pm$ 4.694 & 0.4893 $\pm$ 0.0019 \\
\textbf{\boldmath ASCT-Uniform} & \textbf{\boldmath 0.5283 $\pm$ 0.0022} & \textbf{\boldmath 197.077 $\pm$ 0.024} & \textbf{\boldmath 0.4972 $\pm$ 0.0022} \\
\textbf{\boldmath ASCT-UCT} & \textbf{\boldmath 0.5515 $\pm$ 0.0031} & \textbf{\boldmath 197.588 $\pm$ 0.068} & \textbf{\boldmath 0.5203 $\pm$ 0.0031} \\
\textbf{\boldmath ASCT-AgentUCT} & \textbf{\boldmath 0.5526 $\pm$ 0.0016} & \textbf{\boldmath 193.088 $\pm$ 0.404} & \textbf{\boldmath 0.5220 $\pm$ 0.0016} \\
\bottomrule
\end{tabularx}
\end{table}

	\textbf{Value after amortizing auxiliary work.} Figure~\ref{fig:amortization} projects final-policy value under Equation~\ref{eq:jdeploy}. PPO leads at 100,000 uses; AgentUCT leads at 500,000 and 1,000,000. AgentUCT has higher mean utility and lower auxiliary tokens than VinePPO, giving a mean advantage for every positive $N$. Its mean curve crosses PPO at approximately 360,358 uses. These thresholds assume the specified token weight and subsequent-use utility; only auxiliary expenditure is amortized.
	
	\subsection{RQ3: Does the learned policy transfer across question distributions?}
	\label{sec:transfer}
	\textbf{New question distributions.} Without further updates, we evaluate on 100 unseen 2WikiMultihopQA bridge-comparison questions and 100 MuSiQue questions \citep{ho2020twowiki,trivedi2022musique}. MuSiQue contains 34 two-hop, 33 three-hop, and 33 four-hop items. Supplied passages are mapped to the same RAG environment; policies share questions, frozen components, and alias-aware answer scoring.
	\begin{table}[!htb]
		\centering
		\small
		\setlength{\tabcolsep}{2.8pt}
		\caption{Cross-dataset results on the fixed 100-question cohorts. F1, workflow utility, and execution words are mean \(\pm\) sample SD over three seeds; Base is deterministic.}
		\label{tab:transfer}
		\begin{tabularx}{\linewidth}{@{}Y Y r r r@{}}
			\toprule
			\textbf{Dataset} & \textbf{Method} & \textbf{F1} & \textbf{Utility} & \textbf{Execution words} \\
			\midrule
			2Wiki & Base & 0.5583 & 0.4732 & 3,488.0 \\
			2Wiki & PPO & 0.5883 \(\pm\) 0.0100 & 0.5067 \(\pm\) 0.0102 & 3,342.6 \(\pm\) 130.9 \\
			2Wiki & VinePPO & 0.6517 \(\pm\) 0.0375 & 0.5767 \(\pm\) 0.0360 & 3,069.3 \(\pm\) 133.7 \\
			\textbf{\boldmath 2Wiki} & \textbf{\boldmath ASCT-Uniform} & \textbf{\boldmath 0.6567 \(\pm\) 0.0321} & \textbf{\boldmath 0.5832 \(\pm\) 0.0329} & \textbf{\boldmath 3,009.9 \(\pm\) 47.6} \\
			\textbf{\boldmath 2Wiki} & \textbf{\boldmath ASCT-UCT} & \textbf{\boldmath 0.6583 \(\pm\) 0.0522} & \textbf{\boldmath 0.5837 \(\pm\) 0.0508} & \textbf{\boldmath 3,055.2 \(\pm\) 59.8} \\
			\textbf{\boldmath 2Wiki} & \asctagent{} & \textbf{\boldmath 0.6700 \(\pm\) 0.0321} & \textbf{\boldmath 0.5982 \(\pm\) 0.0313} & \textbf{\boldmath 2,939.3 \(\pm\) 73.7} \\
			MuSiQue & Base & 0.2198 & -0.0215 & 9,880.9 \\
			MuSiQue & PPO & 0.2181 \(\pm\) 0.0029 & -0.0184 \(\pm\) 0.0061 & 9,686.6 \(\pm\) 144.0 \\
			MuSiQue & VinePPO & 0.2248 \(\pm\) 0.0100 & -0.0047 \(\pm\) 0.0128 & 9,397.0 \(\pm\) 156.5 \\
			\textbf{\boldmath MuSiQue} & \textbf{\boldmath ASCT-Uniform} & \textbf{\boldmath 0.2192 \(\pm\) 0.0195} & \textbf{\boldmath -0.0072 \(\pm\) 0.0232} & \textbf{\boldmath 9,274.5 \(\pm\) 181.5} \\
			\textbf{\boldmath MuSiQue} & \textbf{\boldmath ASCT-UCT} & \textbf{\boldmath 0.2225 \(\pm\) 0.0084} & \textbf{\boldmath -0.0041 \(\pm\) 0.0085} & \textbf{\boldmath 9,284.5 \(\pm\) 272.0} \\
			\textbf{\boldmath MuSiQue} & \asctagent{} & \textbf{\boldmath 0.2192 \(\pm\) 0.0135} & \textbf{\boldmath 0.0048 \(\pm\) 0.0082} & \textbf{\boldmath 8,782.3 \(\pm\) 358.6} \\
			\bottomrule
		\end{tabularx}
	\end{table}
	
	Transfer differs by dataset (Table~\ref{tab:transfer}). On 2Wiki, AgentUCT improves both F1 (0.6700 versus 0.6517) and execution words (2,939.3 versus 3,069.3) over VinePPO, raising utility from 0.5767 to 0.5982. On MuSiQue, its lower execution cost offsets slightly lower F1, yielding utility 0.0048 versus $-0.0047$. These unchanged actors use the same component library under new questions and evidence.
	
	\subsection{RQ4: Do trained policies retain description-guided component choice?}
	\label{sec:component}
	\textbf{A newly available component.} A fixed retriever A extends Qwen embedding retrieval with a query instruction and metadata rule. In a 100-question constructed probe, annotation-derived tags mark answer-bearing passages and A prioritizes them. Its implementation, name, corpus, actor weights, and all existing component descriptions stay fixed. Four conditions vary only A's description: type plus the correct evidence relation, type and construction details only, name only, and type plus the opposite relation. Appendix~\ref{app:mechanism} summarizes the protocol.
	\begin{table}[!htb]
		\centering
		\small
		\setlength{\tabcolsep}{2.8pt}
		\caption{Selection rate for the same new retriever under four descriptions. The correct relation states a preference for tagged, answer-bearing evidence; the opposite relation states a preference for untagged passages. Values are mean \(\pm\) sample SD over three seeds.}
		\label{tab:component}
		\begin{tabularx}{\linewidth}{@{}Y r r r r@{}}
			\toprule
			\textbf{Planner} & \shortstack{\textbf{Type +}\\\textbf{correct relation}} & \textbf{Type only} & \textbf{Name only} & \shortstack{\textbf{Type +}\\\textbf{opposite relation}} \\
			\midrule
			Base & 0.820 & 0.010 & 0.000 & 0.020 \\
			PPO & 0.830 \(\pm\) 0.044 & 0.033 \(\pm\) 0.042 & 0.000 \(\pm\) 0.000 & 0.020 \(\pm\) 0.000 \\
			VinePPO & 0.947 \(\pm\) 0.035 & 0.280 \(\pm\) 0.108 & 0.000 \(\pm\) 0.000 & 0.040 \(\pm\) 0.010 \\
			\textbf{\boldmath ASCT-Uniform} & \textbf{\boldmath 0.960 \(\pm\) 0.017} & \textbf{\boldmath 0.300 \(\pm\) 0.111} & \textbf{\boldmath 0.000 \(\pm\) 0.000} & \textbf{\boldmath 0.037 \(\pm\) 0.006} \\
			\textbf{\boldmath ASCT-UCT} & \textbf{\boldmath 0.933 \(\pm\) 0.021} & \textbf{\boldmath 0.197 \(\pm\) 0.068} & \textbf{\boldmath 0.000 \(\pm\) 0.000} & \textbf{\boldmath 0.027 \(\pm\) 0.012} \\
			\asctagent{} & \textbf{\boldmath 0.950 \(\pm\) 0.036} & \textbf{\boldmath 0.373 \(\pm\) 0.205} & \textbf{\boldmath 0.000 \(\pm\) 0.000} & \textbf{\boldmath 0.050 \(\pm\) 0.020} \\
			\bottomrule
		\end{tabularx}
	\end{table}
	
	Selection responds strongly to the description while the component itself stays fixed (Table~\ref{tab:component}). Under the correct evidence relation, ASCT selects A on 93.3--96.0\% of questions, compared with 94.7\% for VinePPO and 82.0\% for Base. Removing the relation lowers ASCT selection to 19.7--37.3\%; name-only descriptions yield zero selection, and the opposite relation yields 2.7--5.0\%. The contrasts show that the trained actors retain description-guided choice across this library extension. This is a controlled selection probe with constructed tags; the cross-dataset study separately evaluates answer quality and execution efficiency.
	
	\FloatBarrier
	\section{Discussion}
	\label{sec:discussion}
	\textbf{What the learned-policy results establish.} The common gain across Uniform, UCT, and AgentUCT supports ASCT's action-evaluation interface across search rules. Uniform's effectiveness is evidence that the benefit does not require cost-aware allocation. The methodological contribution is the connection from executable multi-step alternatives to actor-trajectory updates; the following controls characterize how evaluation supplies that connection.
	
\begin{figure}[!htb]
		\centering
		\includegraphics[width=\linewidth]{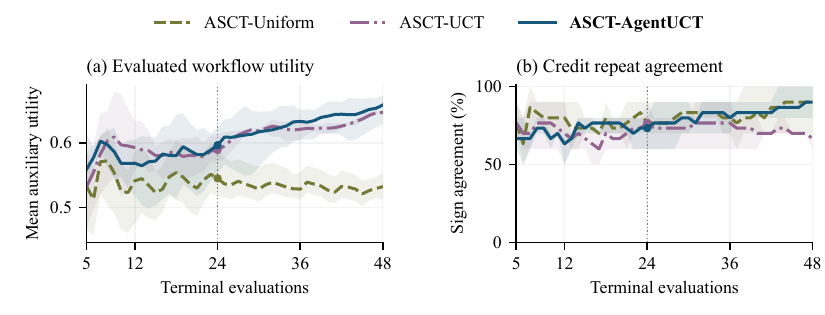}
		\caption{Fixed-state evaluation on two questions, three frozen actor seeds, and two repeats per state. Lines are seed means; bands span seed means. Credit agreement compares signs across repeats. All budgets from 5 to 48 are retained; the guide marks 24.}
		\label{fig:budget_credit}
	\end{figure}

	\textbf{Tree allocation and evaluation budget.} On common five-action states from two questions and three frozen actors, root MC and Uniform share root coverage, uniform suffix sampling, and WTB reuse. At $B=12$, Uniform lowers repeat Q SD from $0.1203$ to $0.0619$ and raises sign agreement from $63.3\%$ to $80.0\%$, using $1.7\%$ more tokens (Table~\ref{tab:root_mc}). The Q-SD ordering reverses at $B=48$. Figure~\ref{fig:budget_credit} traces information acquisition: at 48 trials, UCT, AgentUCT, and Uniform reach evaluated utilities $0.6464$, $0.6584$, and $0.5322$, with sign agreement $66.7\%$, $90.0\%$, and $90.0\%$. These are repeatability diagnostics; RQ1 tests learned-policy effectiveness. Appendix~\ref{app:frozen} supplies costs and full controls.
	
	\textbf{Actor-rollout evaluation within the interface.} With shared actor-centered credit and PPO settings, Uniform improves test utility over ActorRollout at every seed and uses $60.2\%$ fewer auxiliary tokens (Table~\ref{tab:continuation}). ActorRollout balances roots and samples frozen-actor suffixes; this contrast tests the full suffix-evaluation mechanism, including allocation and continuation sampling.
	
	\begin{table}[!htb]
		\centering
		\small
		\setlength{\tabcolsep}{3.2pt}
		\caption{ASCT continuation control: shared $B=12$, credit, PPO settings, and 400 test QIDs. Full-run auxiliary tokens; mean $\pm$ sample SD over three seeds.}
		\label{tab:continuation}
		\begin{tabularx}{\linewidth}{@{}Y r r r@{}}
			\toprule
			\textbf{Continuation variant} & \textbf{Test F1} & \textbf{Test Utility} & \textbf{Aux. tokens (M)} \\
			\midrule
			ASCT-ActorRollout & $0.7142\pm0.0090$ & $0.5975\pm0.0128$ & $495.24\pm13.29$ \\
			ASCT-Uniform & $0.7316\pm0.0168$ & $0.6146\pm0.0191$ & $197.08\pm0.02$ \\
			\bottomrule
		\end{tabularx}
	\end{table}
	
	\textbf{Learning from branch collections.} Our AT$^2$PO adaptation performs close to PPO under the shared configuration (Appendix~\ref{app:mechanism}). Branch records stay fixed within each update and refresh next iteration. This result concerns that adaptation without method-specific tuning.
	
	\section{Limitations}
	\label{sec:limitations}
	Evidence covers one backbone, a finite RAG library, three training seeds, 100-question transfer cohorts, and a constructed description probe. Recoverable execution and finite legal actions delimit applicability. Frozen-state repeatability diagnoses evaluation; it does not isolate each component's contribution to policy gains. Cost projections use the stated auxiliary-token convention; time and price require calibration.
	
	\section{Conclusion}
	\label{sec:conclusion}
	ASCT turns multi-step workflow comparisons into actor-centered credit for PPO on actor-sampled decisions. Three evaluators improve mean held-out utility; cost accounting and transfer studies characterize these gains. The learned policy executes without deployment search.

	\label{main:end}
	
	\section*{Reproducibility statement}
	The experiment package retains per-question results, QID manifests, training ledgers, compact credit records, and fixed-state search traces. Appendices~\ref{app:config}--\ref{app:evidence} document their scope and link reported quantities to the retained artifacts. Paper sources and vector figures accompany the manuscript. Recomputing retained results and reproducing training are separate procedures; training additionally requires the recorded models, datasets, and runtime dependencies.
	
	\section*{AI use statement}
	Following the authors' directions, generative AI tools assisted with (1) identifying related work, (2) implementing experimental methods and code, (3) drafting manuscript sections and explanatory figures, and (4) revising wording. They also provided feedback on experimental methodology and interpretation of results, and assisted with reference, formatting, and code-test checks. The authors take responsibility for the final submission and its claims.

	\FloatBarrier
	\begingroup
	\microtypesetup{protrusion=false}
	\bibliography{references}
	\bibliographystyle{plainnat}
	\endgroup
	
	\appendix
	
	\section{Configuration and execution details}
	\label{app:config}
	\subsection{Evaluator and update implementation}
Search attention specifies how $B$ trials are assigned to root actions and subsequent branches. \textbf{Uniform} selects the least-visited child at fully expanded nodes and samples expansions and continuations uniformly. \textbf{UCT} directs repeat visits using mean utility and an exploration bonus, retaining uniform expansion and continuation. \textbf{AgentUCT} also accounts for predicted uncached continuation cost. At a fully expanded node $v$, its selection score is
\begin{equation}
S(v,a)=\overline U(v,a)
+c_{\mathrm{exp}}\sqrt{\frac{\log(n(v)+1)}{n(v,a)}}
-c_{\mathrm{tok}}\widehat T(v,a),
\label{eq:selection}
\end{equation}
where $\overline U(v,a)$ is the backed-up empirical mean, $n$ denotes visit counts, and $\widehat T$ predicts uncached continuation work. The coefficients control exploration and estimated cost. Expansion and continuation sample candidate actions with probability proportional to $\exp[-c_{\mathrm{tok}}\widehat T(v,a)]$. The implementation averages predicted remaining cost across legal terminal suffixes. These inherited search rules determine the information acquired for Equations~\ref{eq:baseline}--\ref{eq:advantage}.

WTB restores shared materialized prefixes and executes only uncached suffixes, allowing counterfactual comparisons to reuse prior workflow computation. AgentUCT additionally uses cache-conditioned cost estimates to allocate trials; all ASCT evaluators and VinePPO share the reuse infrastructure. A partial hit restores the longest cached prefix and executes the remaining suffix. A terminal hit contributes a logical evaluation with zero new environment tokens. Each method, seed, and worker has an isolated cache namespace. The realized ledger measures executed work; the estimated cost guides allocation.

\subsection{Online PPO training and tree-free deployment}
Each iteration collects fresh actor trajectories under $\pi_{\mathrm{old}}$, constructs their auxiliary credit, and updates the actor. Training standardizes the collected advantages within the iteration. With standardized credit $\widetilde A_E$ and ratio $\rho_\theta=\pi_\theta(a_t\mid s)/\pi_{\mathrm{old}}(a_t\mid s)$, PPO maximizes \citep{schulman2017ppo}
\begin{equation}
L(\theta)=\mathbb E\!\left[
\min\!\left(\rho_\theta\widetilde A_E,
\operatorname{clip}(\rho_\theta,1-\epsilon,1+\epsilon)\widetilde A_E\right)
\right].
\label{eq:ppo}
\end{equation}
The same frozen actor defines action sampling, baseline weights, and the PPO denominator. Thus $\rho_\theta=\pi_\theta(a_t\mid s)/p_{a_t}$ starts at one before updating, avoiding an initial actor--search behavior mismatch. PPO uses actor-sampled trajectories with evaluator-derived credit.

The next iteration evaluates the updated actor's visited states. Online here refers to this alternating collection-and-update procedure. Deployment selects actions from the trained actor and executes the workflow; auxiliary trees are used during training credit construction.

\begin{algorithm}[!tb]
\caption{ASCT with actor-centered credit and PPO}
\label{alg:asct}
\small
\begin{tabularx}{\linewidth}{@{}r@{\quad}X@{}}
\multicolumn{2}{@{}l@{}}{\textbf{Inputs:} actor $\pi_\theta$, training tasks $\mathcal D$, evaluator $E$, budget $B$, utility $U$.}\\
1 & \textbf{for} each training iteration \textbf{do}\\
2 & \quad Freeze $\pi_{\mathrm{old}}\leftarrow\pi_\theta$; initialize actor-record buffer $\mathcal R\leftarrow\varnothing$.\\
3 & \quad \textbf{for} each task in $\mathcal D$ \textbf{do}\\
4 & \qquad Initialize workflow state $s$.\\
5 & \qquad \textbf{while} $s$ is nonterminal \textbf{do}\\
6 & \qquad\quad Compute $p_a=\pi_{\mathrm{old}}(a\mid s)$ for $a\in\mathcal A(s)$; sample $a_t\sim p$.\\
7 & \qquad\quad Initialize an auxiliary tree rooted at a recoverable copy of $s$.\\
8 & \qquad\quad Run $B$ tree trials under $E$, covering all legal root actions first;\\
9 & \qquad\quad after each trial, back up terminal $U(w)$ and log realized auxiliary tokens.\\
10 & \qquad\quad Form $\{\Qhat(s,a)\}_{a\in\mathcal A(s)}$ from the per-root trial means.\\
11 & \qquad\quad Set $\widehat V_E(s)=\sum_a p_a\Qhat(s,a)$ and $\widehat A_E(s,a_t)=\Qhat(s,a_t)-\widehat V_E(s)$.\\
12 & \qquad\quad Append $(s,a_t,p_{a_t},\widehat A_E(s,a_t))$ to $\mathcal R$; advance $s$ using $a_t$.\\
13 & \qquad \textbf{end while}\\
14 & \quad \textbf{end for}\\
15 & \quad Standardize advantages in $\mathcal R$; update $\theta$ with the PPO objective (Eq.~\ref{eq:ppo}).\\
16 & \textbf{end for}; \textbf{return} $\pi_\theta$.\\
\end{tabularx}
\par\smallskip
Auxiliary trials use an independent random stream. Policy updates use the sampled actor records in $\mathcal R$. Deployment executes the trained actor.
\end{algorithm}

	The workflow stages and legal choices appear in Table~\ref{tab:grammar}. The following tables specify the common learner and credit estimators.
	
	The main-task data are mutually disjoint subsets of the official HotpotQA training partition: 2,000 training, 400 validation, and 400 test questions. Both held-out sets contain 76 easy, 256 medium, and 68 hard questions to preserve the training-pool difficulty proportions. Final iteration-3 checkpoints are evaluated with greedy actor actions, using the official answer scorer. We report means and sample standard deviations over seeds 11, 23, and 37.
	
	\textbf{Answer scoring.} Training computes multiset token-overlap F1 after case folding and extracting lowercase alphanumeric tokens; articles are retained and there is no special yes/no/noanswer rule. Held-out HotpotQA reporting uses the official scorer: lowercase, punctuation deletion, article removal, whitespace normalization, and zero credit for a mismatched yes/no/noanswer answer. The reward definition used during collection is fixed across methods; official endpoint scores evaluate the resulting answers under the same reporting rule.
	
	\begin{table}[H]
		\centering
		\small
		\caption{Shared training and evaluation configuration. Training uses the repository answer scorer; endpoint reporting uses the corresponding evaluation scorer.}
		\label{tab:config}
		\begin{tabularx}{\linewidth}{@{}>{\raggedright\arraybackslash}p{0.30\linewidth} Y@{}}
			\toprule
			\textbf{Setting} & \textbf{Value} \\
			\midrule
			Planner & Qwen3-4B-Instruct-2507 \\
			Adapter & LoRA rank 4, alpha 8, dropout 0; query and value projections \\
			Optimizer & AdamW; learning rate \(10^{-5}\) \\
			Optimizer batch size & 8 \\
			PPO & Two update epochs per iteration; clipping \(\epsilon=0.2\) \\
			Advantage processing & Standardization across collected decisions within each iteration \\
			Train / validation / test & 2,000 / 400 / 400 mutually disjoint QIDs \\
			Held-out difficulty counts & 76 easy, 256 medium, 68 hard in each set \\
			Seeds and iterations & 11, 23, 37; three iterations per trained method \\
			Endpoint rule & Final iteration-3 checkpoint; greedy actor decisions \\
			Auxiliary budget & \(B=12\) terminal trials per actor-visited nonterminal state \\
			Workflow utility & \(\lambda=0.1\), \(C_0=4096\) execution words \\
			UCT exploration & \(c_{\mathrm{exp}}=1.4\) \\
			AgentUCT allocation cost & \(c_{\mathrm{tok}}=10^{-4}\); cache-conditioned suffix estimate \\
			Cost reporting & \(\beta=10^{-4}\) per auxiliary token: environment execution and auxiliary actor scoring \\
			\bottomrule
		\end{tabularx}
	\end{table}
	
	\begin{table}[H]
		\centering
		\small
		\caption{Experimental methods. All trained rows use the same planner family and PPO update settings.}
		\label{tab:methods}
		\begin{tabularx}{\linewidth}{@{}Y Y Y@{}}
			\toprule
			\textbf{Method} & \textbf{Credit at a sampled actor decision} & \textbf{Auxiliary evaluation} \\
			\midrule
			Base & No policy update & None \\
			PPO & Terminal trajectory utility & None \\
			VinePPO & Adjacent Monte Carlo state values & Actor continuations \\
			\textbf{\boldmath ASCT-Uniform} & \textbf{\boldmath Actor-centered action advantage} & \textbf{\boldmath Uniform tree evaluator} \\
			\textbf{\boldmath ASCT-UCT} & \textbf{\boldmath Actor-centered action advantage} & \textbf{\boldmath Utility-directed UCT evaluator} \\
			\asctagent{} & \textbf{\boldmath Actor-centered action advantage} & \textbf{\boldmath Cost-aware AgentUCT evaluator} \\
			\bottomrule
		\end{tabularx}
	\end{table}
	
	The planner scores legal action labels and normalizes their probabilities over the current legal set. Collection samples actions from this distribution; evaluation takes its maximum. Advantages are standardized over each iteration's collected decisions before PPO updates. Auxiliary samples update the estimator attached to those decisions.
	
	The workflow execution counter uses lowercase alphanumeric word units. Retrieval charges corpus words for each query; reciprocal-rank fusion charges its four constituent rankings. Maximal marginal relevance and model reranking add the work defined by their passage sets. Generation charges the selected answer-context words. This common execution proxy is distinct from tokenizer-based model-call counts.
	
	\textbf{Workflow-adapted VinePPO.} At each actor-visited nonterminal state, the frozen actor supplies $B=12$ independently sampled continuations with replacement; their mean terminal utility is the stored state value. Credit is $r_t+\widehat V(s_{t+1})-\widehat V(s_t)$ with unit discount. Intermediate rewards are zero; the final actor transition receives its workflow utility and has successor value zero. The stored estimate at the next actor-visited state supplies the preceding transition's successor value, without an additional value-estimation call. Thus $B$ counts auxiliary terminal samples per visited nonterminal state, not per side of a value difference. Main actor execution is separate. Cached terminal outcomes still count as logical samples; cached probabilities are reused, but action draws are fresh. WTB reuse and actor-scoring costs follow the declared ledger.
	
	\textbf{Legal-action probabilities.} For each legal label, the planner scores its JSON action response under the same prompt. The score is the mean conditional log probability over response tokens (excluding prompt and padding tokens). A softmax over these length-normalized scores defines the categorical action distribution. Collection samples this distribution, and PPO uses the corresponding stored normalized action log probability; greedy evaluation selects its maximum. This scoring rule is shared by the compared actors.
	
	\subsection{Fixed prompts and query refinement}
\label{app:prompts}
Table~\ref{tab:prompts} gives the core instructions used across methods. The actor prompt is fixed while its LoRA weights are trained; query and answer generation disable the adapter. The templates are implemented in \nolinkurl{code/main/src/atppo_lab/policy.py}. The table omits shared JSON-format reminders and retry messages, which remain in the released code.

\begin{table}[H]
\centering\small
\caption{Fixed LLM instructions and their runtime inputs. Quoted instructions are taken from the implementation; output constraints are summarized.}
\label{tab:prompts}
\begin{tabularx}{\linewidth}{@{}>{\raggedright\arraybackslash}p{.17\linewidth} Y >{\raggedright\arraybackslash}p{.25\linewidth}@{}}
\toprule
\textbf{Role / input} & \textbf{Core instruction} & \textbf{Output constraint} \\
\midrule
Actor / current workflow observation & ``You select the next action in a RAG workflow.'' & JSON with one \texttt{action} field; probabilities normalized over legal labels. \\
Query2Doc / question & ``Write a short hypothetical passage that would answer the question.'' & JSON with one \texttt{pseudo\_document} string, used as the retrieval query. \\
Decomposition / question & ``Decompose the multi-hop question into exactly two focused search queries.'' & JSON \texttt{queries}: exactly two nonempty strings. \\
Answer / question and evidence passages & ``Answer only from the evidence. Return the shortest answer span that satisfies the question.'' & JSON \texttt{answer} and \texttt{citations}; use \texttt{UNKNOWN} if evidence is insufficient and cite relevant evidence. \\
\bottomrule
\end{tabularx}
\end{table}

\textbf{Deterministic refinement.} Continue uses \texttt{BridgeQueryRefiner} in \nolinkurl{code/main/src/atppo_lab/rag.py}, with no LLM call. It prioritizes evidence whose title occurs in the question, extracts a bridge entity with fixed text patterns, and falls back to an unseen evidence title, then the first title or the original question. Fixed templates target birth, death, government position, or location; otherwise the query is ``Facts about \{entity\} needed to answer: \{question\}.'' Fixed workflow rules implement the original-question action, Stop/Continue limits, retrieval widths, and context cutoffs.

	\section{Token ledger and cost-adjusted value}
	\label{app:tokens}
	
	The main-text auxiliary-token ledger separates environment execution and auxiliary actor scoring:
	\begin{equation}
		T_{\mathrm{aux}}=T_{\mathrm{env}}+T_{\mathrm{actor,aux}}.
	\end{equation}
	The environment term expands as
	\begin{equation}
		T_{\mathrm{env}}=T_{\mathrm{4B,in}}+T_{\mathrm{4B,out}}
		+T_{\mathrm{query\ embedding,0.6B}}+T_{\mathrm{reranker,0.6B}}.
	\end{equation}
	These are realized counts after reuse. Auxiliary actor scoring charges the recorded tokens for evaluating continuation-action probabilities. Main-trajectory actor scoring, optimizer forward passes, E5, BGE-M3, and shared index construction have separate counters. AgentUCT's allocation estimate is a cache-conditioned proxy for remaining generative work; the reporting ledger includes the recorded embedding and reranking calls. All token categories use unit weight under the stated $\beta$.
	
	\begin{table}[H]
		\centering
		\small
		\setlength{\tabcolsep}{3.4pt}
		\caption{Auxiliary evaluation cost within ASCT. Totals are millions of environment tokens. Per-state values normalize within each seed. Search return charges realized auxiliary work as in Equation~\ref{eq:jsearch}. Values are mean $\pm$ sample SD.}
		\label{tab:cost}
		\begin{tabularx}{\linewidth}{@{}Y r r r@{}}
			\toprule
			Evaluator & Total tokens (M) & Tokens / actor state & $\overline J_{\mathrm{search}}$ \\
			\midrule
			ASCT-Uniform & 197.077 $\pm$ 0.024 & 3,732.5 $\pm$ 3.2 & 0.4972 $\pm$ 0.0022 \\
			ASCT-UCT & 197.588 $\pm$ 0.068 & 3,749.5 $\pm$ 2.6 & 0.5203 $\pm$ 0.0031 \\
			\asctagent{} & \textbf{\boldmath 193.088 $\pm$ 0.404} & \textbf{\boldmath 3,664.5 $\pm$ 9.2} & \textbf{\boldmath 0.5220 $\pm$ 0.0016} \\
			\bottomrule
		\end{tabularx}
	\end{table}
	
\begin{figure}[p]
		\centering
		\includegraphics[width=\linewidth,height=.72\textheight,keepaspectratio]{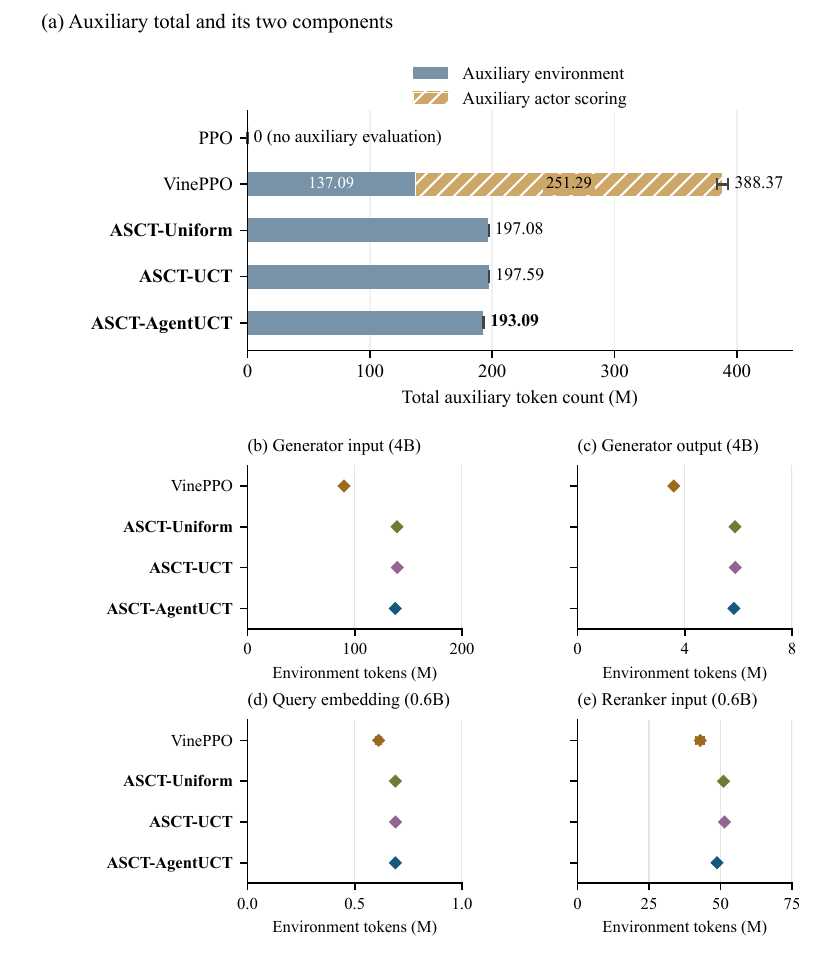}
		\caption{Auxiliary token volume for the recorded Qwen model operations over complete three-iteration training runs. (a) Stacked means sum environment calls and auxiliary actor scoring; endpoint whiskers give the sample SD of each seed's summed count. Numbers at the right are totals. (b--e) Environment categories for VinePPO and the three ASCT evaluators, in the same top-to-bottom order. Each panel has an independent zero-based token axis. Diamonds and whiskers show mean $\pm$ sample SD across three seeds. M denotes one million tokens. Boldface identifies all ASCT variants.}
		\label{fig:token_categories}
	\end{figure}

	Table~\ref{tab:token_ledger} reports the combined ledger in Section~\ref{sec:rq3}. Figure~\ref{fig:token_categories} shows total auxiliary volume first, followed by the four environment categories for VinePPO and ASCT. PPO has zero auxiliary evaluation tokens; its main-trajectory and optimizer scoring remain part of training. The lower panels use a common method order and an independently labeled, zero-based scale for each category.
	
	For method $m$ and seed $r$, the projection is computed before aggregation:
	\begin{equation}
		J_{\mathrm{deploy},m,r}(N)=U_{m,r}^{\mathrm{test}}-\beta T_{m,r}^{\mathrm{aux}}/N.
	\end{equation}
	Reported standard deviations are across these seed-level projections. For a comparison with positive mean utility difference $\Delta\overline U$ and positive mean token difference $\Delta\overline T$, the crossing of mean curves is
	\begin{equation}
		N_* = \beta\Delta\overline T/\Delta\overline U.
	\end{equation}
	With total auxiliary cost, \asctagent{} versus VinePPO has $\Delta\overline U=0.02481849$ and $\Delta\overline T=-195{,}286{,}252.33$ tokens. Higher mean utility and lower mean cost give a positive projected value difference for every $N>0$.
	
	\begin{table}[H]
		\centering
		\small
		\setlength{\tabcolsep}{3.4pt}
		\caption{Projected final-policy value after amortizing total auxiliary tokens: $J_{\mathrm{deploy}}(N)$, mean $\pm$ sample SD over three seed-level projections. $N$ is the assumed number of uses. These values correspond to Figure~\ref{fig:amortization}.}
		\label{tab:amortization}
		\begin{tabularx}{\linewidth}{@{}Y r r r@{}}
			\toprule
			Method & $N=100{,}000$ & $N=500{,}000$ & $N=1{,}000{,}000$ \\
			\midrule
			PPO & 0.5651 $\pm$ 0.0021 & 0.5651 $\pm$ 0.0021 & 0.5651 $\pm$ 0.0021 \\
			VinePPO & 0.2055 $\pm$ 0.0046 & 0.5162 $\pm$ 0.0056 & 0.5550 $\pm$ 0.0059 \\
			\textbf{\boldmath ASCT-Uniform} & \textbf{\boldmath 0.4175 $\pm$ 0.0190} & \textbf{\boldmath 0.5752 $\pm$ 0.0191} & \textbf{\boldmath 0.5949 $\pm$ 0.0191} \\
			\textbf{\boldmath ASCT-UCT} & \textbf{\boldmath 0.4099 $\pm$ 0.0241} & \textbf{\boldmath 0.5680 $\pm$ 0.0241} & \textbf{\boldmath 0.5878 $\pm$ 0.0241} \\
			\asctagent{} & \textbf{\boldmath 0.4256 $\pm$ 0.0075} & \textbf{\boldmath 0.5801 $\pm$ 0.0076} & \textbf{\boldmath 0.5994 $\pm$ 0.0076} \\
			\bottomrule
		\end{tabularx}
	\end{table}
	
	\FloatBarrier
	
	\section{Difficulty-resolved policy results}
	\label{app:learning_slices}
	
	The test questions are grouped by their supplied HotpotQA difficulty labels. Each method--seed group contains the same 76 easy, 256 medium, and 68 hard questions. Utility uses the official answer scorer and the execution-word penalty in Equation~\ref{eq:utility}. We compute each seed's group mean before reporting the mean and sample SD across seeds.
	
	\begin{table}[!htb]
\centering\small
\caption{Final-policy utility by HotpotQA test difficulty, using the official answer scorer. Values are mean $\pm$ sample SD across three training seeds; Base is deterministic. The three difficulty groups partition the same 400 test questions used in Table~\ref{tab:main}.}
\label{tab:difficulty}
\begin{tabularx}{\linewidth}{@{}Y r r r@{}}
\toprule
Method & Easy ($n=76$) & Medium ($n=256$) & Hard ($n=68$) \\
\midrule
Base & 0.6348 & 0.5723 & 0.4582 \\
PPO & 0.6405 $\pm$ 0.0023 & 0.5702 $\pm$ 0.0050 & 0.4616 $\pm$ 0.0053 \\
VinePPO & 0.6613 $\pm$ 0.0046 & 0.6031 $\pm$ 0.0108 & 0.4836 $\pm$ 0.0191 \\
\textbf{\boldmath ASCT-Uniform} & \textbf{\boldmath 0.6565 $\pm$ 0.0045} & \textbf{\boldmath 0.6334 $\pm$ 0.0265} & \textbf{\boldmath 0.4968 $\pm$ 0.0097} \\
\textbf{\boldmath ASCT-UCT} & \textbf{\boldmath 0.6540 $\pm$ 0.0056} & \textbf{\boldmath 0.6222 $\pm$ 0.0345} & \textbf{\boldmath 0.5005 $\pm$ 0.0097} \\
\textbf{\boldmath ASCT-AgentUCT} & \textbf{\boldmath 0.6583 $\pm$ 0.0111} & \textbf{\boldmath 0.6338 $\pm$ 0.0178} & \textbf{\boldmath 0.5174 $\pm$ 0.0120} \\
\bottomrule
\end{tabularx}
\end{table}

	\section{Learning interfaces, evaluator rules, and paired records}
	\label{app:paired}
\begin{figure}[!htb]
\centering
\includegraphics[width=\linewidth]{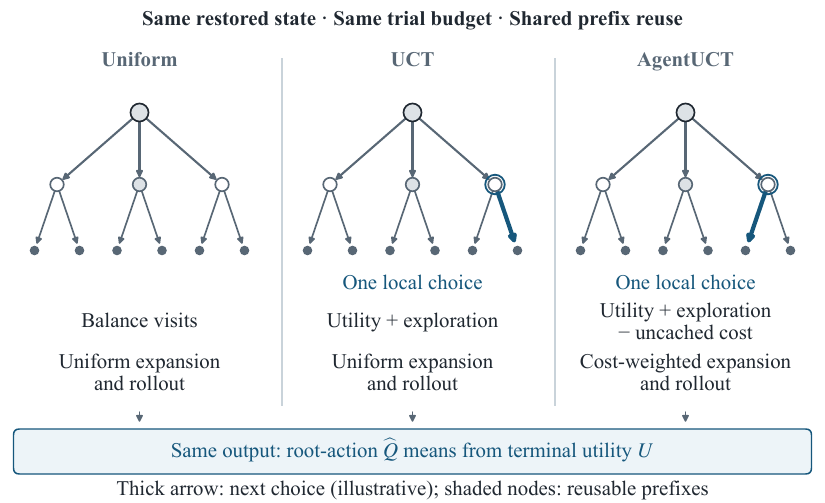}
\caption{Decision rules within the three ASCT evaluators. UCT and AgentUCT share the depicted tree and prefix; the thick arrow differs only at one circled internal node, illustrating a possible local change in the next choice. Arrow thickness does not encode visit counts or the frequency or magnitude of cost effects. AgentUCT adds predicted uncached cost to selection and uses cost-weighted sampling. All three use WTB prefix reuse, back up terminal utility $U$, and return root-action means \citep{li2026agentuct}.}
\label{fig:evaluator_flow}
\end{figure}

	Figure~\ref{fig:training_interface} in Section~\ref{sec:method} shows how evaluation enters learning. Figure~\ref{fig:evaluator_flow} details the allocation rules.

	Seed-level endpoints and paired utility contrasts are retained in the accompanying data files, including \texttt{paired\_seed\_differences.csv}.
	
	The three training seeds are the units for variability reporting. Methods share the same test questions, enabling paired contrasts. The reported sample standard deviations describe observed variability; the smaller evaluator differences have mixed signs across seeds.
	
	\FloatBarrier
	\section{Training credit and frozen-state allocation controls}
\label{app:frozen}

\textbf{Credit on the observed actor trajectories.} We analyze the existing ASCT-Uniform training records for all three seeds: 2,000 training questions across three iterations give 6,000 trajectories per seed and 158,402 decision records in total. Each record contains the sampled action, frozen actor probabilities, legal-action Q table and raw advantage. We reconstruct the actor-weighted baseline and sampled-action advantage from these fields. Keeping the recorded Q table fixed, we also compute the sampled action's advantage under a uniform-action baseline. Table~\ref{tab:observed_credit} reports the resulting sign differences on states with at least three legal actions. Both compared advantages must have magnitude above $10^{-8}$ to count a sign change.

\begin{table}[!htb]
\centering\small
\caption{Baseline semantics on actual ASCT-Uniform training decisions. Counts cover three training iterations per seed. Multi-action states have at least three legal actions. Sign changes compare actor-weighted and uniform-action centering of the same recorded Q table at the actual sampled action.}
\label{tab:observed_credit}
\begin{tabular}{@{}rrrrr@{}}
\toprule
Seed & All decisions & Multi-action decisions & Sign changes & Rate (\%) \\
\midrule
11 & 52,828 & 12,000 & 1,043 & 8.69 \\
23 & 52,833 & 12,000 & 1,021 & 8.51 \\
37 & 52,741 & 12,000 & 1,053 & 8.78 \\
\bottomrule
\end{tabular}
\end{table}

Across the three seeds, 3,117 of 36,000 multi-action decisions change sign (8.66\%). These are actions that actually entered training, extending the mechanism analysis beyond selected counterfactual examples. The calculation identifies the local comparison supplied by actor-centered credit. Final learning outcomes are evaluated by the complete-method training comparisons in Section~\ref{sec:rq1}; the alternative baseline here is an offline recomputation.

\textbf{Root-stratified controls on common five-action states.} The fixed-state follow-up uses the original training split's question indices 5 and 6. Each completed iteration-3 ASCT-Uniform actor (seeds 11, 23, and 37) supplies its sampled retriever state, with five legal root actions. Evaluators share its materialized prefix, actor probabilities and sampled action. Every run starts with a fresh WTB namespace and reset transient ranking/reranking caches; WTB reuse remains enabled within the run. Prefix reconstruction and index preparation precede measurement. Each evaluator has two repeats of 48 terminal trials, retaining every intermediate Q table and executed token increment. The 12-trial comparison uses the corresponding prefixes.

Root MC balances root trials and draws each subsequent legal action independently and uniformly. Uniform tree evaluation uses uniform expansion and rollout sampling but balances visits at internal nodes. This comparison tests the allocation structure under the same suffix-sampling rules and cache availability. Internal allocation changes which suffixes are evaluated; it need not estimate the same fixed continuation distribution as independent MC.

\begin{table}[!htb]
\centering\small
\setlength{\tabcolsep}{3pt}
\caption{Internal tree allocation at the training budget ($B=12$). Both evaluators cover legal root actions, use uniform expansion/rollout sampling, and retain WTB reuse. Two training questions provide five-action retriever states for each of three frozen actors; two repeats per state. Cells report mean $\pm$ sample SD across seed means. Q SD and credit-sign agreement measure repeatability.}
\label{tab:root_mc}
\begin{tabularx}{\linewidth}{@{}Y r r r@{}}
\toprule
Evaluator & Repeat Q SD & Credit agreement (\%) & Tokens (k) \\
\midrule
Root-stratified MC & $0.1203\pm0.0653$ & $63.3\pm11.5$ & $11.09\pm0.13$ \\
ASCT-Uniform & $0.0619\pm0.0293$ & $80.0\pm10.0$ & $11.28\pm0.51$ \\
\bottomrule
\end{tabularx}
\end{table}

\begin{table}[!htb]
\centering\small
\setlength{\tabcolsep}{3pt}
\caption{Root-stratified MC and Uniform tree evaluation on the same five-action states. All values are mean $\pm$ sample SD across three seed means. Each state has two repeats; budgets are prefixes of the completed 48-trial records.}
\label{tab:root_mc_budget}
\begin{tabularx}{\linewidth}{@{}r Y r r r@{}}
\toprule
$B$ & Evaluator & Repeat Q SD & Credit (\%) & Tokens (k) \\
\midrule
12 & Root-stratified MC & $0.1203\pm0.0653$ & $63.3\pm11.5$ & $11.09\pm0.13$ \\
12 & ASCT-Uniform & $0.0619\pm0.0293$ & $80.0\pm10.0$ & $11.28\pm0.51$ \\
48 & Root-stratified MC & $0.0368\pm0.0088$ & $83.3\pm5.8$ & $38.15\pm1.03$ \\
48 & ASCT-Uniform & $0.0484\pm0.0158$ & $90.0\pm17.3$ & $36.98\pm1.67$ \\
\bottomrule
\end{tabularx}
\end{table}

At 12 trials, repeat Q SD decreases by 48.6\% and credit-sign repeat agreement increases by 16.7 percentage points with internal allocation, at 1.7\% greater executed token cost. At 48 trials, Uniform's Q SD is 0.0484 versus root MC's 0.0368, while sign agreement is 90.0\% versus 83.3\%. The observed repeatability benefit applies to the small budget used in training; the ordering depends on budget. Q SD measures variation across repeated evaluations and credit agreement measures sign repeatability; neither uses an accuracy target.

\textbf{Four-question binary-state corroboration.} A separate completed diagnostic uses the first four training questions and the first control and reranker state visited by each of the same three actors: eight binary-action states per seed, 24 checkpoint--state combinations in total. Five evaluators each make four repeats of 12 trials. Root MC balances root trials with uniform suffixes; Actor MC balances roots with frozen-actor suffixes. Two independent 96-trial Actor-MC batches provide a pooled actor-continuation reference. The resulting 528 evaluations retain complete Q tables, terminal paths, utilities and cost counters.

\begin{table}[!htb]
\centering\small
\setlength{\tabcolsep}{3pt}
\caption{Four-question binary-state diagnostic: eight states per actor seed, four repeats at $B=12$. Order agreement uses two independent 96-trial actor-continuation batches as its reference. Total auxiliary tokens include actor scoring. Values are mean $\pm$ sample SD across three seed means.}
\label{tab:binary_probe}
\begin{tabularx}{\linewidth}{@{}Y r r r@{}}
\toprule
Evaluator & Repeat Q SD & Ref. order (\%) & Tokens (k) \\
\midrule
Root-stratified MC & $0.0149\pm0.0221$ & $97.9\pm3.6$ & $3.90\pm0.31$ \\
ASCT-Uniform & $0.0064\pm0.0093$ & $99.0\pm1.8$ & $4.16\pm0.31$ \\
ASCT-UCT & $0.0093\pm0.0148$ & $96.9\pm5.4$ & $4.18\pm0.32$ \\
ASCT-AgentUCT & $0.0085\pm0.0121$ & $99.0\pm1.8$ & $4.07\pm0.37$ \\
Actor MC & $0.0038\pm0.0040$ & $99.0\pm1.8$ & $4.78\pm0.53$ \\
\bottomrule
\end{tabularx}
\end{table}

Uniform tree evaluation reduces repeat Q SD relative to root MC in every actor seed, from 0.0149 to 0.0064 on average, using 4,158 versus 3,899 environment tokens. The order-agreement ceiling reflects this binary cohort: both reference batches agree on the action order at every state. For two actions, any nondegenerate convex baseline lies between their Q values; comparing actor and uniform baselines therefore cannot reveal a sign difference. The actual multi-action training decisions above supply the relevant centering evidence.

\textbf{Functional cost-phase controls.} On the five-action follow-up states, eight settings independently enable cost awareness in selection, expansion and rollout. The all-off setting is UCT and the all-on setting is native AgentUCT. With Uniform and root MC as controls, the design contains 120 runs and 5,760 terminal trials. For each state, a separate 240-leaf enumeration weights terminal paths by the product of uniform legal-action probabilities, producing a reference for that named continuation distribution. Six enumerations add 1,440 reference terminals. The earlier one-question pilot retains 30 runs and three further enumerations in the evidence package.

\begin{table}[!htb]
\centering\small
\setlength{\tabcolsep}{3pt}
\caption{Cost-aware selection, expansion and rollout (S/E/R) on the same two questions and three actors. Bits enable cost awareness at each phase; 000 is UCT and 111 is native AgentUCT. Cells are mean $\pm$ sample SD across seed means, with two repeats per state. WTB reuse is enabled throughout.}
\label{tab:cost_phases}
\begin{tabularx}{\linewidth}{@{}Y r r r r@{}}
\toprule
S/E/R & Tokens@12 (k) & Tokens@48 (k) & Q SD@48 & Credit@48 (\%) \\
\midrule
UCT (000) & $11.14\pm0.63$ & $35.66\pm1.36$ & $0.0609\pm0.0150$ & $66.7\pm5.8$ \\
001 & $10.94\pm0.66$ & $35.93\pm0.73$ & $0.0528\pm0.0128$ & $80.0\pm20.0$ \\
010 & $11.69\pm1.55$ & $37.01\pm1.96$ & $0.0526\pm0.0123$ & $76.7\pm15.3$ \\
011 & $11.55\pm1.29$ & $34.90\pm2.24$ & $0.0492\pm0.0219$ & $80.0\pm10.0$ \\
100 & $11.24\pm0.57$ & $34.94\pm2.12$ & $0.0471\pm0.0197$ & $70.0\pm17.3$ \\
101 & $11.01\pm0.50$ & $36.15\pm0.68$ & $0.0450\pm0.0074$ & $76.7\pm5.8$ \\
110 & $11.77\pm1.52$ & $36.29\pm1.16$ & $0.0519\pm0.0029$ & $73.3\pm23.1$ \\
AgentUCT (111) & $11.56\pm1.47$ & $34.46\pm0.75$ & $0.0503\pm0.0239$ & $90.0\pm10.0$ \\
Uniform & $11.28\pm0.51$ & $36.98\pm1.67$ & $0.0484\pm0.0158$ & $90.0\pm17.3$ \\
Root MC & $11.09\pm0.13$ & $38.15\pm1.03$ & $0.0368\pm0.0088$ & $83.3\pm5.8$ \\
\bottomrule
\end{tabularx}
\end{table}

At 48 trials, full AgentUCT reduces mean executed tokens by 3.4\% relative to UCT and improves credit-sign repeat agreement from 66.7\% to 90.0\%. At 12 trials its token count is 3.7\% higher. The switches therefore affect the executed search and its local comparisons, with effects that depend on phase combinations and budget. All per-trial records are retained, including reference discrepancies and selection changes. The reference describes uniform-continuation values; discrepancies for adaptive evaluators are reference alignment, not universal Q error. These controls make no policy updates.

\textbf{Budget curves from completed evaluations.} The original training records use 12 terminal evaluations per state and contain endpoint action tables. For longer budget curves, we use the completed 48-trial fixed-state evaluations on two questions already in the same training split (ordered indices 5 and 6). Each of the three iteration-3 actors supplies its observed five-action retriever state. Two repeated searches per method retain every terminal utility, cumulative executed token count, and reconstructed action table. Prefixes of these records provide every budget from 5 to 48, including 24, with no additional environment execution or policy updates. The same materialized prefix and actor probabilities are used across evaluators within each state. Fresh per-evaluation WTB namespaces and reset transient retrieval/reranker caches isolate auxiliary execution; prefix preparation is outside the measured ledger.

\begin{figure}[!htb]
\centering
\includegraphics[width=\linewidth]{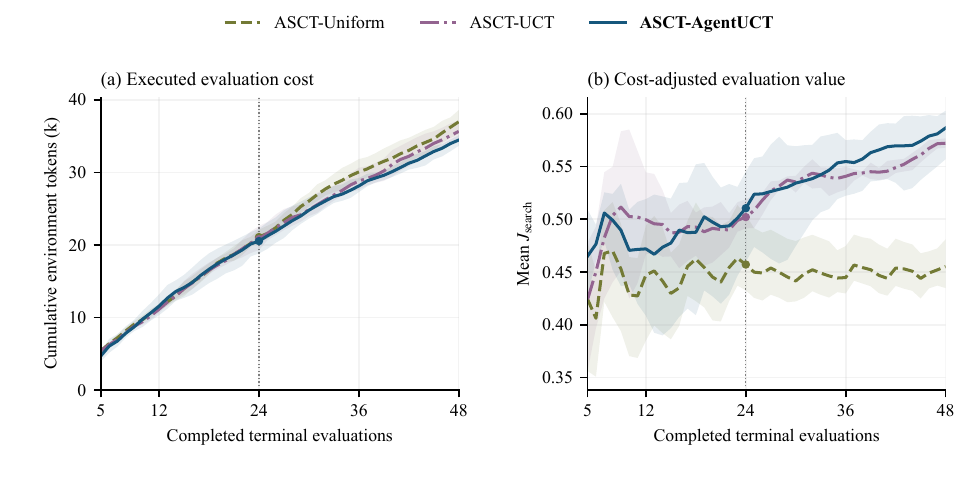}
\caption{Executed auxiliary tokens and cost-adjusted evaluation value on the same states and budgets as Figure~\ref{fig:budget_credit}. $\overline J_{\mathrm{search}}(K)=\overline U(K)-10^{-4}T_{\mathrm{aux}}(K)/K$. Auxiliary actor scoring is zero for these three evaluators, so total auxiliary cost equals environment cost. The deployment amortization curve uses Equation~\ref{eq:jdeploy}. Lines and bands follow Figure~\ref{fig:budget_credit}.}
\label{fig:budget_cost}
\end{figure}

At budget $K$, mean evaluated utility is $K^{-1}\sum_{i=1}^{K}U(w_i)$. Credit repeat agreement is the fraction of legal actions whose centered-credit signs agree between the two independent searches at that budget. It measures repeatability, with final-policy utility assessed separately in the training experiments. These ASCT evaluators use rule-based continuations and add no auxiliary actor-scoring calls, so $T_{\mathrm{aux}}(K)=T_{\mathrm{env}}(K)$. Equation~\ref{eq:jsearch} gives $\overline J_{\mathrm{search}}(K)$ using $\beta=10^{-4}$. We average repeats and questions within each seed before reporting seed means. The bands in Figures~\ref{fig:budget_credit} and~\ref{fig:budget_cost} span the three seed-level means; they are descriptive ranges.

\begin{table}[!htb]
\centering\small\setlength{\tabcolsep}{4pt}
\caption{The 24-evaluation checkpoint of the full curves. Values average the same two questions, two repeats and three actor seeds. The complete curves retain seed-level variability.}
\label{tab:budget24}
\begin{tabular}{@{}lrrrr@{}}
\toprule
Evaluator & Mean utility & Tokens & $\overline J_{\mathrm{search}}$ & Credit agreement (\%) \\
\midrule
ASCT-Uniform & 0.5449 & 21,077.9 & 0.4571 & 76.7 \\
ASCT-UCT & 0.5887 & 20,788.5 & 0.5021 & 76.7 \\
\asctagent{} & \textbf{0.5960} & \textbf{20,531.3} & \textbf{0.5105} & \textbf{73.3} \\
\bottomrule
\end{tabular}
\end{table}

\textbf{Observed relation.} At 24 evaluations, \asctagent{} has the highest mean auxiliary utility and $\overline J_{\mathrm{search}}$ among the three evaluators, while its credit repeat agreement is lower (Table~\ref{tab:budget24}). At 48 evaluations, UCT and AgentUCT reach mean utilities 0.6464 and 0.6584, compared with 0.5322 for Uniform. Their corresponding $\overline J_{\mathrm{search}}$ values are 0.5721, 0.5866, and 0.4551. Credit repeat agreement is 66.7\%, 90.0\%, and 90.0\%, respectively. Thus, directing evaluation toward higher-utility continuations and producing repeatable action comparisons are distinct properties. These exploratory curves describe information acquisition on two existing training questions. The main experiments measure the learned policy's quality and the aggregate execution cost across training.

\section{Component protocol and continuation diagnostics}
\label{app:mechanism}
\textbf{Description controls.} Component A prioritizes passages tagged from answer-bearing annotations in a constructed corpus. The four descriptions vary whether its type and evidence preference are stated correctly, only its construction is stated, no description is given, or the opposite preference is stated. Its implementation, name, corpus, actor weights, and existing component descriptions stay fixed. The correct description also explains the tags, so the comparison does not isolate one wording feature. Per-question answer outcomes are retained in the accompanying data package.

\textbf{What makes the credit useful?} For a fixed table, the local surrogate $L_s(\theta)=\sum_a\pi_\theta(a\mid s)\Qhat(s,a)$ has logit derivative $p_a\widehat A_E(s,a)$ at the behavior policy. This explains the direction of the raw credit before normalization and clipping. It benefits actor utility when evaluator preferences align with outcomes the actor can realize. The continuation rule determines which downstream outcomes contribute to each action value. Online recollection refreshes visited states and actor-weighted baselines after each training iteration.

\textbf{Continuation control.} ASCT-Uniform uses uniform tree expansion and continuation with least-visited selection; ASCT-ActorRollout balances root actions and samples suffix actions from the frozen actor. Both retain root coverage and actor-centered credit. This intervention changes the full continuation mechanism, and each condition trains its own evolving actor. Table~\ref{tab:continuation} reports the three-seed held-out endpoints. The matched-QID ASCT-ActorRollout minus ASCT-Uniform test-utility differences are $-0.0077$, $-0.0294$, and $-0.0143$ in seeds 11, 23, and 37. All 18,000 Actor-CF training trajectories were retained; reconstructing the Q-table centering for 158,357 actor decisions found no protocol violations. The audit verifies that recorded credits match the action values and actor-weighted baselines.

Across complete training runs, ASCT-ActorRollout uses $495.24\pm13.29$ million auxiliary tokens: $180.12$ million in environment execution and $315.12$ million in auxiliary actor scoring on average. Main-trajectory and optimizer actor scoring are excluded, as in the shared ledger. Its mean auxiliary rollout utility is $0.5488$.

\textbf{AT$^2$PO workflow adaptation.} We adapted the public tree learner \citep{zong2026at2po} to categorical workflow actions under the planned training setup, with batch size eight and three collect--update iterations. Each iteration collects trees with the current actor and optimizes over that collection; the next iteration collects again with the updated actor. Each training question produces a 106-leaf tree whose branch records enter optimization. The training trajectories remain fixed during each update and are refreshed at the next collection.

Independent iteration-3 evaluation on the paper's test QIDs with the official scorer gives test utility $0.5647\pm0.0042$ and F1 $0.6945\pm0.0041$ across three seeds. Relative to PPO ($0.5651\pm0.0021$), the paired test-utility difference is $-0.0004\pm0.0063$. Auxiliary tokens total $696.15\pm19.53$ million per run; optimizer scoring is separate. This is an exploratory evaluation of one adaptation without a method-specific tuning study. The adaptation differs from ASCT in its evaluation unit, credit construction, and policy-update records.

\section{Retained evidence}
\label{app:evidence}
The accompanying data package retains matched per-QID endpoints, seed-level auxiliary ledgers, compact credit records, and the component and transfer probes. For the new controls, the compact supplement includes three-seed matched-QID predictions, training metrics, and protocol audits. Full Actor-CF and AT$^2$PO trajectory archives are retained separately by the authors. These artifacts support recomputation of reported results; reproducing training additionally requires the recorded models, indices, and runtime.
	
\end{document}